\documentclass[runningheads]{llncs}

\usepackage{appendix}
\usepackage{cite, makecell}
\usepackage{float}
\usepackage{amsmath,amssymb,amsfonts}
\usepackage{algorithmic}
\usepackage{graphicx}
\usepackage{caption}
\graphicspath{{images/}}
\usepackage{textcomp}
\usepackage{xcolor}
\usepackage{balance}
\usepackage{multirow}
\usepackage{hyphenat}

\def\BibTeX{{\rm B\kern-.05em{\sc i\kern-.025em b}\kern-.08em
		T\kern-.1667em\lower.7ex\hbox{E}\kern-.125emX}}
\usepackage{amssymb,amsmath,epsfig,latexsym, bm}
\usepackage{nohyperref}	%pageanchor=false,
\hypersetup{colorlinks, pdfa=true, breaklinks, citecolor=black, urlcolor=black, linkcolor=black}
\usepackage{amssymb,amsmath,epsfig,latexsym, bm}
\usepackage{mathtools, mdwmath, mdwtab, multirow, amssymb, amsmath, array, ragged2e}

\usepackage{threeparttable, booktabs, url, eqparbox, xspace, setspace, tabularx, subfig}
\makeatletter
\DeclareRobustCommand\onedot{\futurelet\@let@token\@onedot}
\def\@onedot{\ifx\@let@token.\else.\null\fi\xspace}

\def\etal{\emph{et al}\onedot}
\makeatother

\begin{document}
\title{DeepTriNet: A Tri-Level Attention Based DeepLabv3+ Architecture for Semantic Segmentation of Satellite Images}

\author{\vspace{.85in}}

 \author{Tareque Bashar Ovi\textsuperscript{1}\orcidID{0000-0001-6961-8894} \and Shakil Mosharrof\textsuperscript{2}\orcidID{0000-0002-7228-2823} \and Nomaiya Bashree\textsuperscript{3}\orcidID{0009-0008-6151-2356} \and Md Shofiqul Islam\textsuperscript{4}\orcidID{0000-0002-2597-4050} \and  Muhammad Nazrul Islam\textsuperscript{5}\orcidID{0000-0002-7189-4879}}

\authorrunning{T. B. Ovi~\etal}
\titlerunning{Satellite Image Segmentation}
 \institute{\textsuperscript{1-5}Military Institute of Science and Technology (MIST)\\
 Mirpur Cantonment, Dhaka--1216, Bangladesh\\
 \textsuperscript{1}\email{ovitareque@gmail.com}\hspace{0.5cm}
 \textsuperscript{2}\email{shakilmrf8@gmail.com}\hspace{0.5cm}
\textsuperscript{3}\email{nomaiyabashree2002@gmail.com}\hspace{0.5cm}
\textsuperscript{4} \email{shafiqcseiu07@gmail.com}\hspace{0.5cm}
 \textsuperscript{5}\email{nazrul@cse.mist.ac.bd}\hspace{0.5cm}
 }

\maketitle              % typeset the header of the contribution
\begin{abstract}
The segmentation of satellite images is crucial in remote sensing applications. Existing methods face challenges in recognizing small-scale objects in satellite images for semantic segmentation primarily due to ignoring the low-level characteristics of the underlying network and due to containing distinct amounts of information by different feature maps.  Thus, in this research, a tri-level attention-based DeepLabv3+ architecture (DeepTriNet) is proposed for the semantic segmentation of satellite images.  The proposed hybrid method combines squeeze-and-excitation networks (SENets) and tri-level attention units (TAUs) with the vanilla DeepLabv3+ architecture, where the TAUs are used to bridge the semantic feature gap among encoders output and the SENets used to put more weight on relevant features. The proposed DeepTriNet finds which features are the more relevant and more generalized way by its self-supervision rather we annotate them. The study showed that the proposed DeepTriNet performs better than many conventional techniques with an accuracy of 98\% and 77\%, IoU 80\% and 58\%, precision 88\% and 68\%, and recall of 79\% and 55\% on the 4-class Land-Cover.ai dataset and the 15-class GID-2 dataset respectively. The proposed method will greatly contribute to natural resource management and change detection in rural and urban regions through efficient and semantic satellite image segmentation
\end{abstract}


\section{Introduction}  
Satellite image segmentation is required to monitor and evaluate land cover and land use. In other words, it is required for the management of natural resources. Different methods are applied to analyse images and data acquired from remote sensing  to offer land description and change detection in rural and urban areas. In many areas, including urban planning, a fundamental task for ecological environment conservation, vegetation monitoring, and even military reconnaissance, detailed information about land use or land cover is an invaluable resource\cite{6032741}, making precise satellite image segmentation a research aspect.

Traditional manual interpretation and classification of satellite images is the process of visually analyzing and categorizing satellite images based on human expertise and visual inspection, which can be time-consuming and error-prone. Again, semantic segmentation involves the implementation of pixel-level labels to images, which is a crucial task\cite{10.2307/37940} in study of satellite images for land cover classification and urban planning.

The use of images in remote sensing (RS) applications has drawn more attention in recent years due to the recent expansion in satellite and aerial imagery availability. Achievements in computer vision, natural language processing, and audio processing have all been made possible by advances in machine learning (ML), larger benchmark datasets, and improved CPU capacity. Deep learning techniques, which have become the standard methods for remote sensing image landcover classification research, automatically extract low-level image features of objects from images by creating deep networks, then combine them into high-level abstract features to achieve higher classification accuracy\cite{article2}. Despite the availability of petabytes of freely available satellite imagery and a wide range of benchmark datasets for various RS tasks, the widespread success and popularity of machine learning—particularly of deep learning methods — has not yet been fully adopted to the RS domain. This is not to argue that there aren't any successful ML applications in RS, but rather that the promise of the nexus between both domains hasn't been entirely fulfilled.

%The fact that geographical imaging datasets are much more diverse in their content than datasets generated for conventional vision applications presents a significant barrier in many RS activities. For instance, most traditional cameras produce 3-channel RGB imagery, whereas most satellite sensors produce various spectral band sets. To solve this issue attention mechanism is generally used extensively which helps a CNN to extract more efficient features from the input image. Deep learning models can incorporate attention mechanisms, a particular kind of neural network layer. By giving varied weights to various input components, they enable the model to concentrate on particular portions of the input. This weighting is often determined by how pertinent each input component is to the task at hand.%
Therefore, the objective of this research is to propose an effective deep-learning network using several attention techniques to reduce the semantic information gap between the encoder and decoder, and for allowing the accurate segmentation of satellite pictures. As outcomes, this research proposed a tri-level attention-based DeepLabv3+ architecture (DeepTriNet) for the semantic segmentation of satellite images.  As such, the contribution could be summarized as: (i) integrating a self-supervised attention mechanism with vanilla DeepLabv3+ that combines channel-level, spatial-level, and pixel-level abstractions for better generalization of the pertinent information.(ii)Introducing Squeeze-and-Excitation(SE) in the decoder portion of the proposed architecture that explicitly models channel inter-dependencies and adaptively re-calibrates channel-wise feature outputs.(iii) Evaluating the robustness of the model by performing experiments on two datasets, one of which has $15$ classes but is comparatively small in volume and the other one with only $4$ classes but large in volume.

\section{Related Works}  
Convolutional Neural Network (CNN) usage in the past years has shown promising results in automating satellite image segmentation tasks.
 Using FastFCN, Onim~\etal~\cite{Onim2020} reported an accuracy of 93\%, precision of 99\%, recall of 0.98, and mIoU of 97\% using the GID-2 dataset. Only $6$ classes were used for their study.
 
 Tong~\etal~\cite{Zhang2020(b)}  proposed CNN based land cover classification algorithm that uses deep learning, hierarchical segmentation, and multi-scale information fusion to classify satellite images using the GID-2 dataset. ResNet-50 was used for initial training purposes. The proposed method outperforms traditional methods such as colour histogram, grey-level co-occurrence matrix, and local binary patterns in land cover classification accuracy.
 
 Yao~\etal\cite{Yao2019} used high-resolution remote sensing photos and two image datasets containing six common land cover classes to classify land use using a deep convolutional neural network method with an attention mechanism. With a Kappa coefficient of 0.91, the suggested model surpassed other models in terms of accuracy and metrics value, achieving an overall accuracy of 93.5\%. 
 
 For LULC use using remote sensing data, a unique combined deep learning framework incorporating MLP and CNN models were implemented by Zhang~\etal\cite{zhang2020}. The framework's accuracy ranged from 85.5\% to 92.3\%, which was higher overall than previous techniques and less susceptible to smaller sample sizes. In order to separate land use and land cover classes from RGB satellite pictures, Nayem~\etal\cite{Nayem2020} developed a semantic segmentation framework utilizing the FCN-8 algorithm. Their proposed architecture had an average intersection over union (IoU) of 84\% and an average accuracy of 91.0\% on the GID-2 dataset. Kang~\etal\cite{KANG2021102499} proposed a Multi-scale context extractor network for water-body extraction from high-resolution satellite images using the LandCover.ai dataset. Though $93.23$\% of mean IOU was reported in their study, only $1$ class(water) was considered for their experiment. Lee~\etal\cite{lee2022comparisons} performed a comparative analysis between various deep learning architectures using the LandCover.ai dataset and reported $88.4$\%, $91.4$\%, and $85.8$\% for SegNet, U-Net, and DeepLabv3+ respectively.

Recent literature shows that there were many attempts to semantically segment satellite images with the GID-2 dataset~\cite{Nayem2020, Onim2020, Yao2019}, using traditional end-to-end deep learning models. On the other hand, few networks were used to segment maps with other LandCover.ai datasets. But self-supervised attention mechanism-based  satellite image segmentation on the GID-2 with $15$ classes and LandCover.ai dataset with  is yet to be tested.

\section{Proposed Architecture}
This section broadly introduces the theoretical concepts that includes
\subsubsection{DeepLabv3+} The purpose of DeepLabv3+ is to assign semantic labels (such as a person, a dog, or a cat) to every pixel in the input image proposed by Chen~\etal~\cite{Liang2017}. The DeepLabv3+ model contains two phases: encoding and decoding. The encoding phase extracts the essential information from the image using a convolutional neural network (CNN) whereas the decoding phase reconstructs the output of appropriate dimensions based on the information obtained from the encoder phase. The decoder module was included to improve object boundary segmentation results. MobileNetv2, Xception, ResNet, PNASNet, and Auto-DeepLab are among the network backbones that Deeplab supports. It is used as a benchmark for semantic segmentation today.

\subsubsection{Tri-Level Attention Unit (TAU)} 
\begin{figure*}[htpb]
    \centering
    \includegraphics[width=1.\columnwidth]{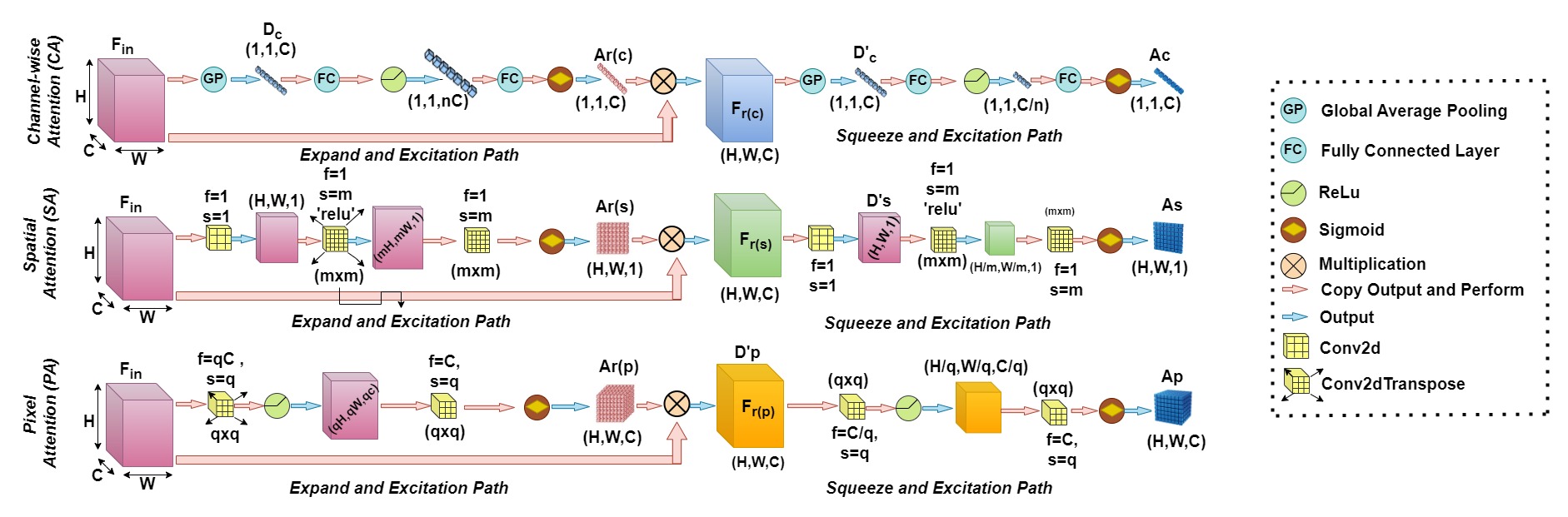}
    \caption{Tri-Level Attention Unit Architecture}
    \label{TAU}
\end{figure*}

\begin{figure*}[htpb]
    \centering
    \includegraphics[width=1.0\columnwidth]{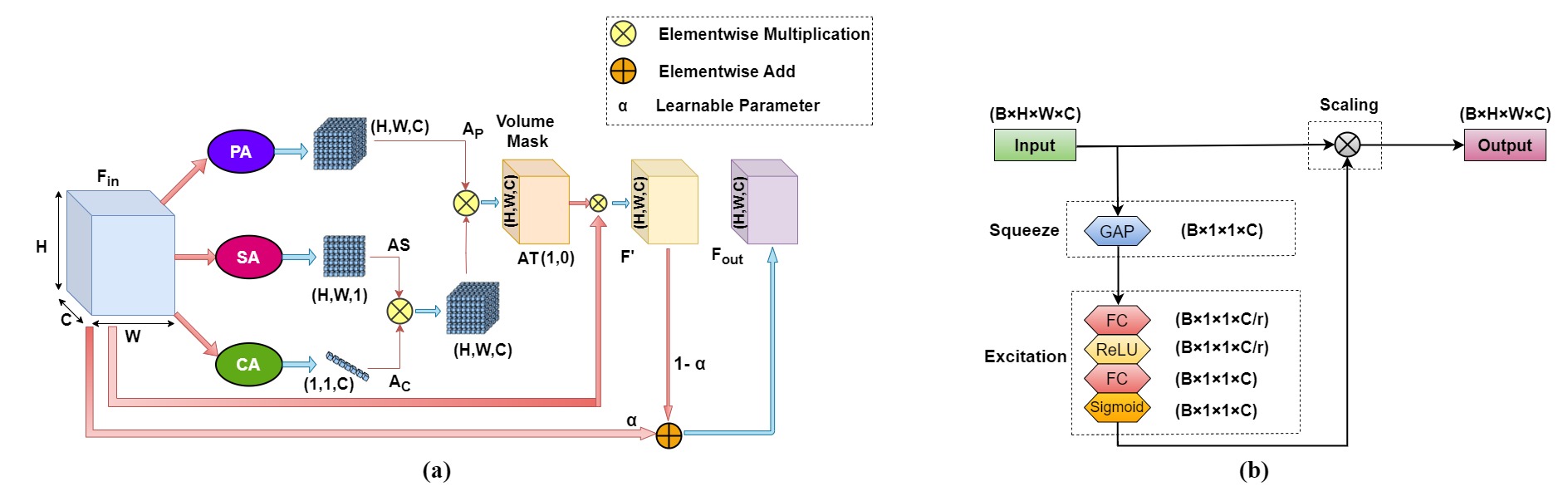}
    \caption{(a)~TAU Mechanism and (b)~SENets Architecture}
    \label{TAU_SE}
\end{figure*}

Tri-Level Attention Unit (TAU)  proposed by Tanvir Mahmud~\etal~\cite{Mahmud_2021}  combines three levels of abstraction—channel, spatial, and pixel—for greater generalization of the pertinent contextual information. It is a novel self-supervised attention technique. The channel attention (CA) mechanism functions from a wider perspective to highlight the corresponding channels that contain more information, the spatial attention (SA) mechanism focuses more on the local spatial regions that contain regions of interest, and finally, the pixel attention (PA) mechanism functions from the lowest level to assess the feature relevance of each pixel. Thus TAU unit module is frequently utilized throughout the DeepTriNet decoder network to recalibrate features and increase feature relevance. TAU architecture and mechanism is given in 
Fig.\ref{TAU} and Fig.\ref{TAU_SE}(a).

\subsubsection{Squeeze-and-Excitation Networks(SENets)}Squeeze-and-Excitation Networks (SENets) provide a CNN building block that enhances channel interdependencies at essentially no computational cost. In CNN Before combining the data over all available output channels, the various filters will first locate spatial features in each input channel. All there is to know for now is that the network equally weights each channel when producing the output feature maps. By including a content-aware system to adaptively weight each channel, SENets aim to change this. In its simplest form, this may entail providing each channel with a single parameter and a linear scalar representing how relevant each one is. By condensing the feature maps to a single numerical value, they first gain a broad comprehension of each channel. As a result, a vector of size n is produced, where n is the number of convolutional channels. It is then input into a two-layer neural network, which produces an output vector of the same size. These n values can now be applied to the original feature maps as weights, sizing each channel according to its significance. This whole mechanism is visualised in Fig.\ref{TAU_SE}(b).
\begin{center}
    \centering
    \includegraphics[height= 80mm,width=87mm]{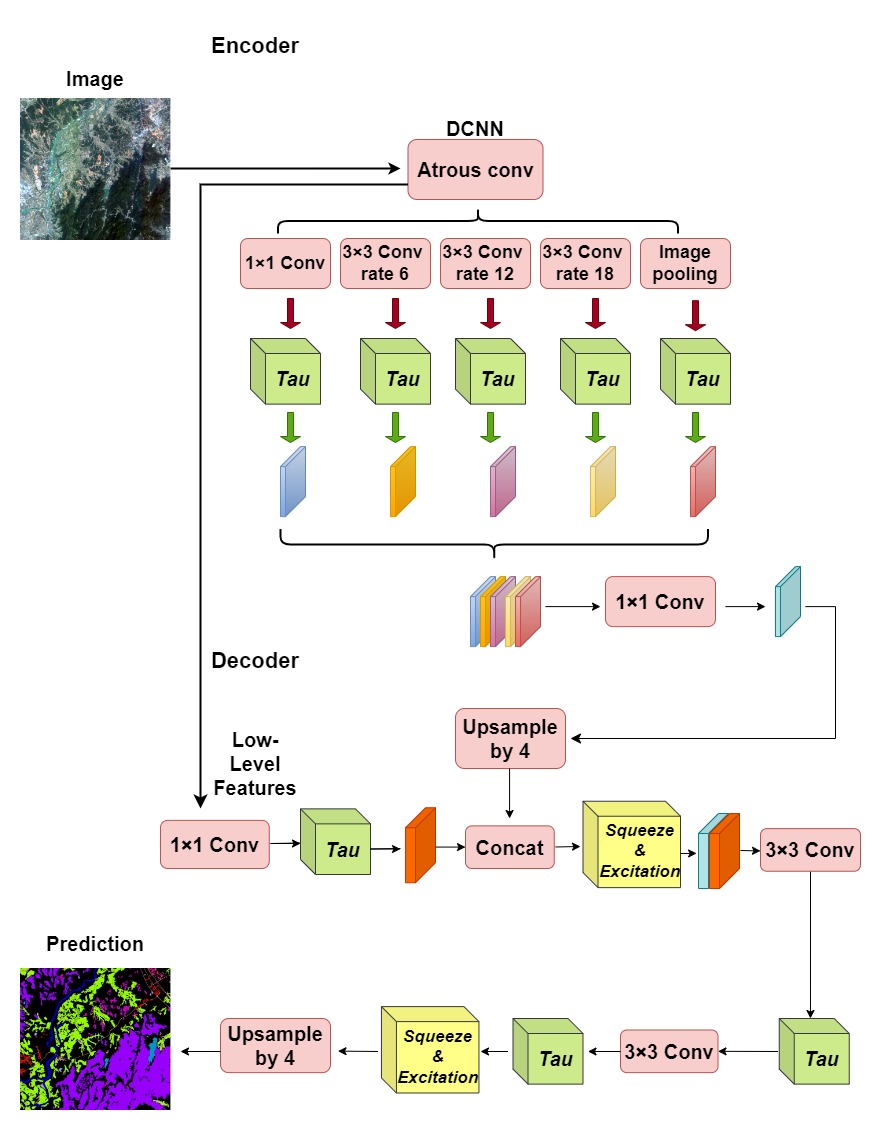}
    \captionof{figure}{Proposed Architecture}
    \label{proposedArch}
\end{center}
\subsubsection{Proposed Architecture } In this research both the Tri-Level Attention Unit (TAU) and the Squeeze-and-Excitation Networks (SENets) have been integrated extensively with the vanilla DeepLabv3+ architecture. More specifically, the TAU has been integrated with the Atrous Spatial Pyramid Pooling(ASPP) portion of the architecture, where each convolution has been passed through the TAU to extract more refined features. Squeeze-and-Excitation Networks (SENets) has been introduced at the decoder portion of this network. Detailed architecture of our proposed network is given in the Fig.\ref{proposedArch}

\section{Dataset}

\subsection{Dataset Description}
\begin{figure*}[htpb]
    \centering
    \includegraphics[width=0.55\columnwidth]{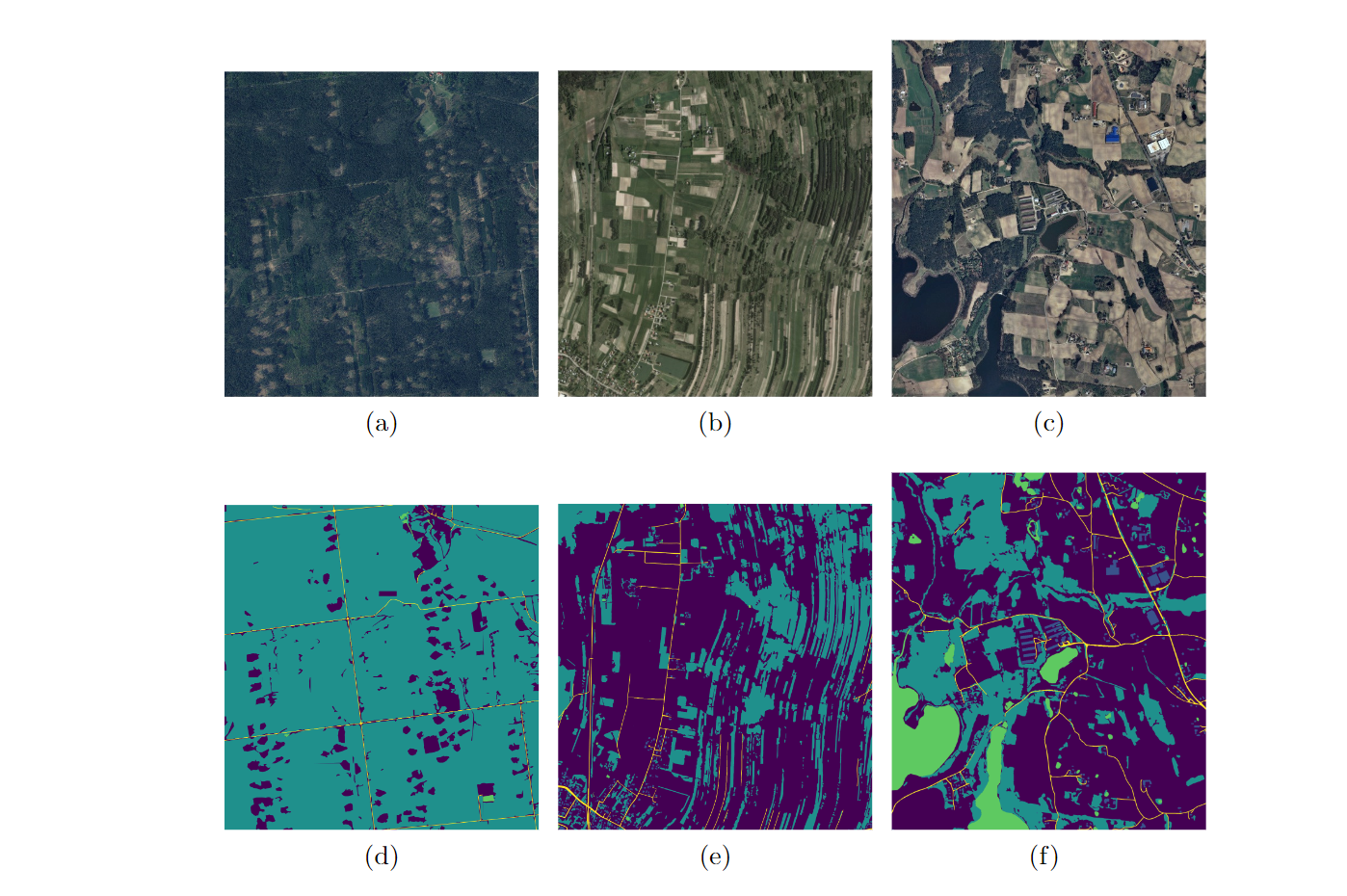}
    \caption{Example of images from the LandCover.ai~\cite{https://doi.org/10.48550/arxiv.2005.02264} dataset: (a--c)~Input image and (d--f)~Target image}
    \label{lc_dataset}
\end{figure*}

Two open-source dataset LandCover.ai~\cite{https://doi.org/10.48550/arxiv.2005.02264} and  Gaofen-2 Image Dataset (GID-2)~\cite{GID2020} are used in this study. LandCover.ai dataset has four classes: building (1), woods (2), water (3), and road (4).
 Again Gaofen-2 Image Dataset (GID-2) has mainly two portions: a large-scale classification set with six classes and a fine land-cover classification set with 15 classes. This work has used the latter part to verify the  proposed architecture with more classes. An example of these two datasets is given in fig.\ref{lc_dataset} and \ref{gid_dataset}.
%\begin{itemize}
%    \item explains the types of data in the dataset
%    \item explain grid and patch and their necessity
%    \item  explains the relationship between target and input images
 %   \item attach some visual images with label information
%\end{itemize}

 \begin{center}
    \centering
    \includegraphics[width=0.55\columnwidth]{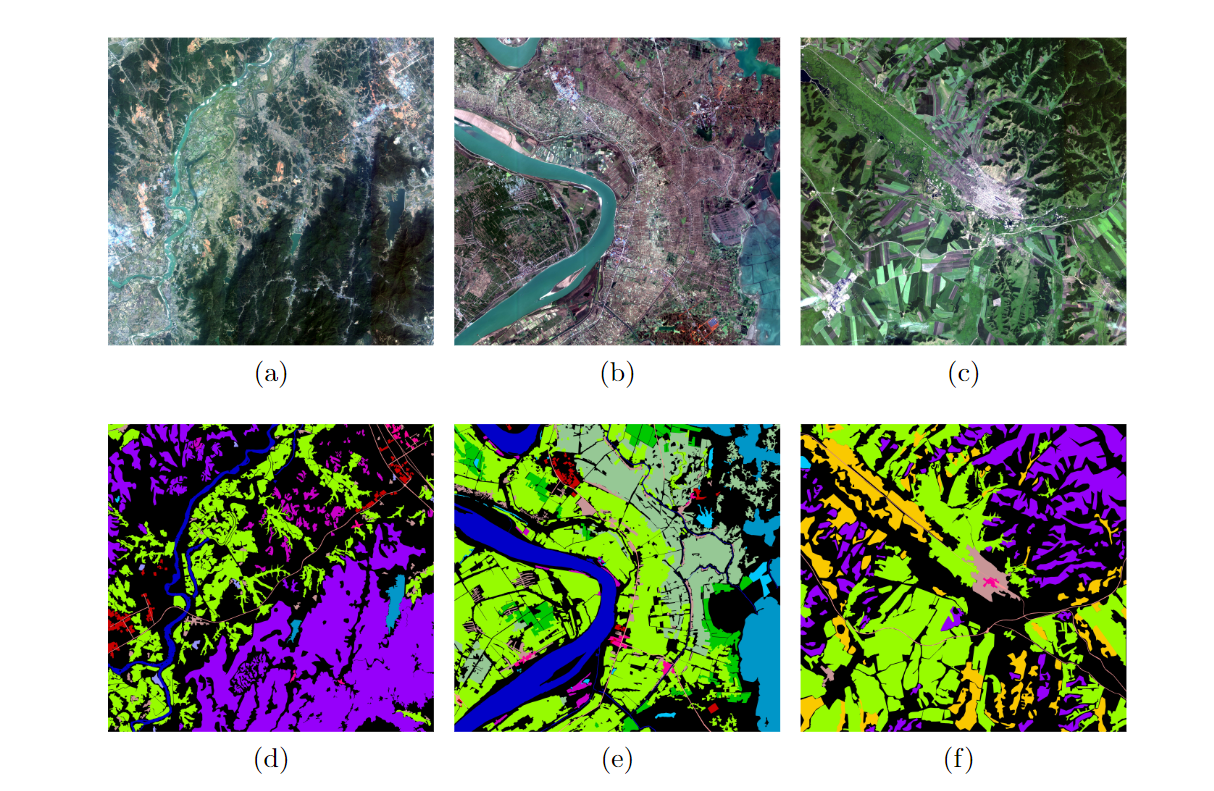}
    \captionof{figure}{Example of images from the Gaofen-2 Image Dataset (GID-2)~\cite{GID2020}: (a--c)~Input image and (d--f)~Target image}
    \label{gid_dataset}
\end{center}

\subsection{Dataset Preprocessing}

\begin{figure*}[htpb]
    \centering
    \includegraphics[width=0.75\columnwidth]{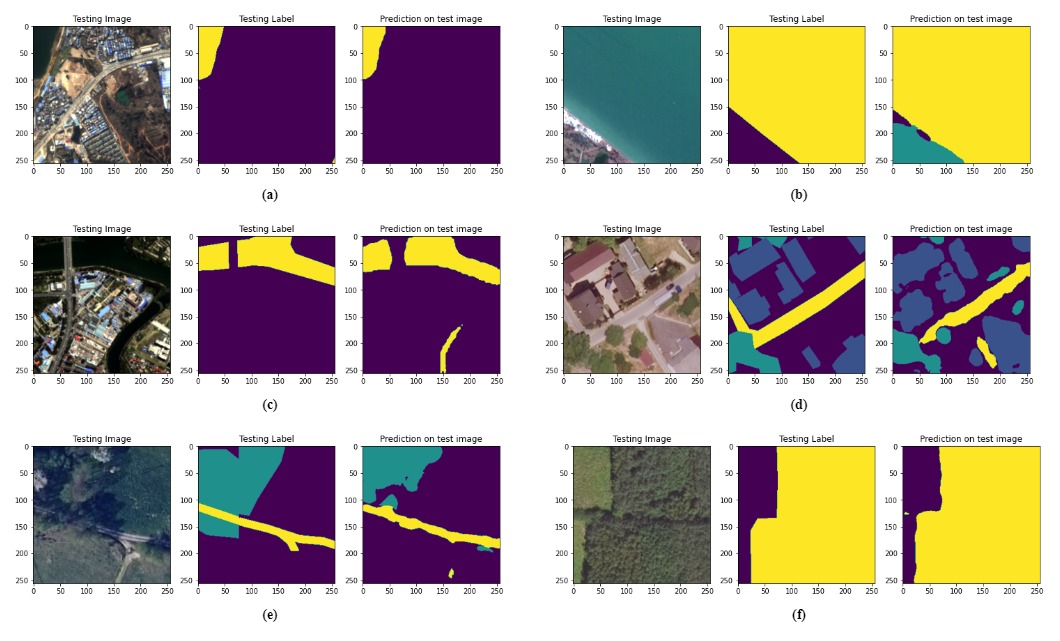}
    \caption{Patch wise Prediction of DeepTriNet on GID-2 Dataset}
    \label{prediction_gid}
\end{figure*}

Tri-Level attention-based DeepLabv3+ requires the input image to be downsampled to $256\times256\times3$. This results in a significant loss of pixel information. To address and solve this issue, we followed the grid and patch method introduced in the work of Onim~\etal~\cite{Onim2020}. Each of the  original and annotated images was gridded to 728 sub-images of shape $256\times256\times3$. The sub-images were bounded to the grid and patch block to preserve their spatial position. Detailed patch-wise prediction of DeepTriNet on GID-2 and LandCover.ai dataset is shown in Fig.\ref{prediction_gid} and \ref{prediction_lc} respectively.

\begin{center}
    \centering
    \includegraphics[width=0.75\columnwidth]{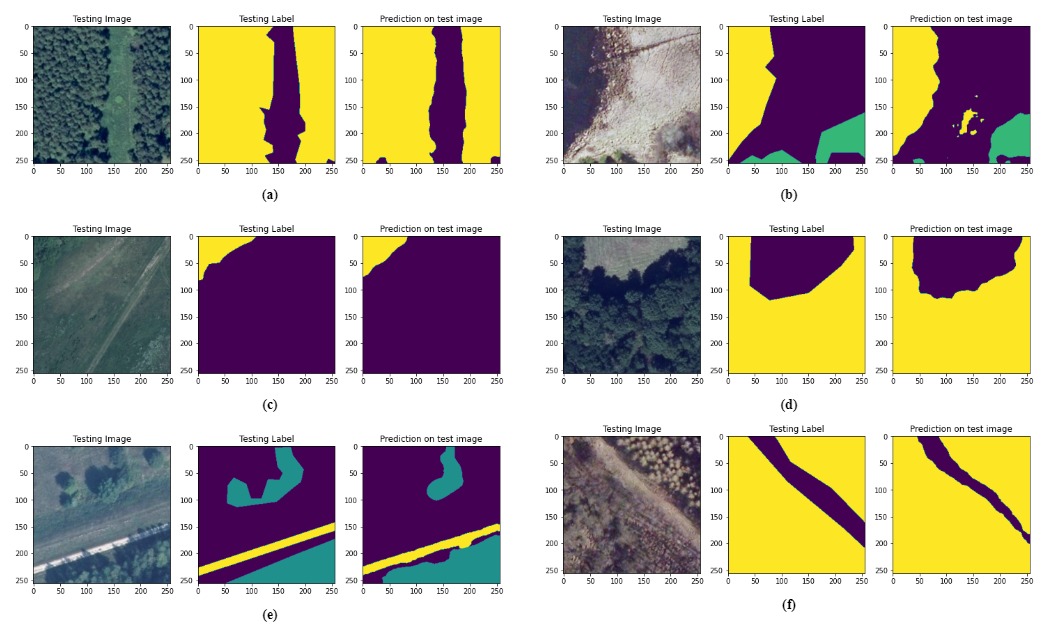}
    \captionof{figure}{Patch-wise Prediction of DeepTriNet on LandCover.ai Dataset}
    \label{prediction_lc}
\end{center} 

The final preprocessing step normalized the pixel values from $-1$ to $1$ as the Equation \eqref{norm}. Here $i, j$ represents image height and width respectively, $I_{i,j}$ is the original image and $I^{n}_{i,j}$ is the normalized image.

\begin{equation}
    \label{norm}
    I^{n}_{i,j} = \frac{I_{i,j}}{127.5}-1
\end{equation}

\section{Methodology}
\subsection{Proposed Methods}
 Our research and evaluation workflow can be explained as follows.
 \begin{itemize}
    \item At first, both the training images and ground truths are subdivided into 784 images each by gridding the high-resolution satellite images.
    \item After that, these patches were normalized and then used for the training of the DeepTriNet.
    \item After the training, the prediction is done for each patch by the best saved weight, and after that smooth tiled prediction were done to reconstruct a high-resolution image. The generation of the evaluation metrics for each images follows  through pixel-by-pixel calculation. The whole process is visualized in Fig.\ref{research_workflow}
\begin{figure*}[htpb]
    \centering
    \includegraphics[height= 70mm,width=100mm]{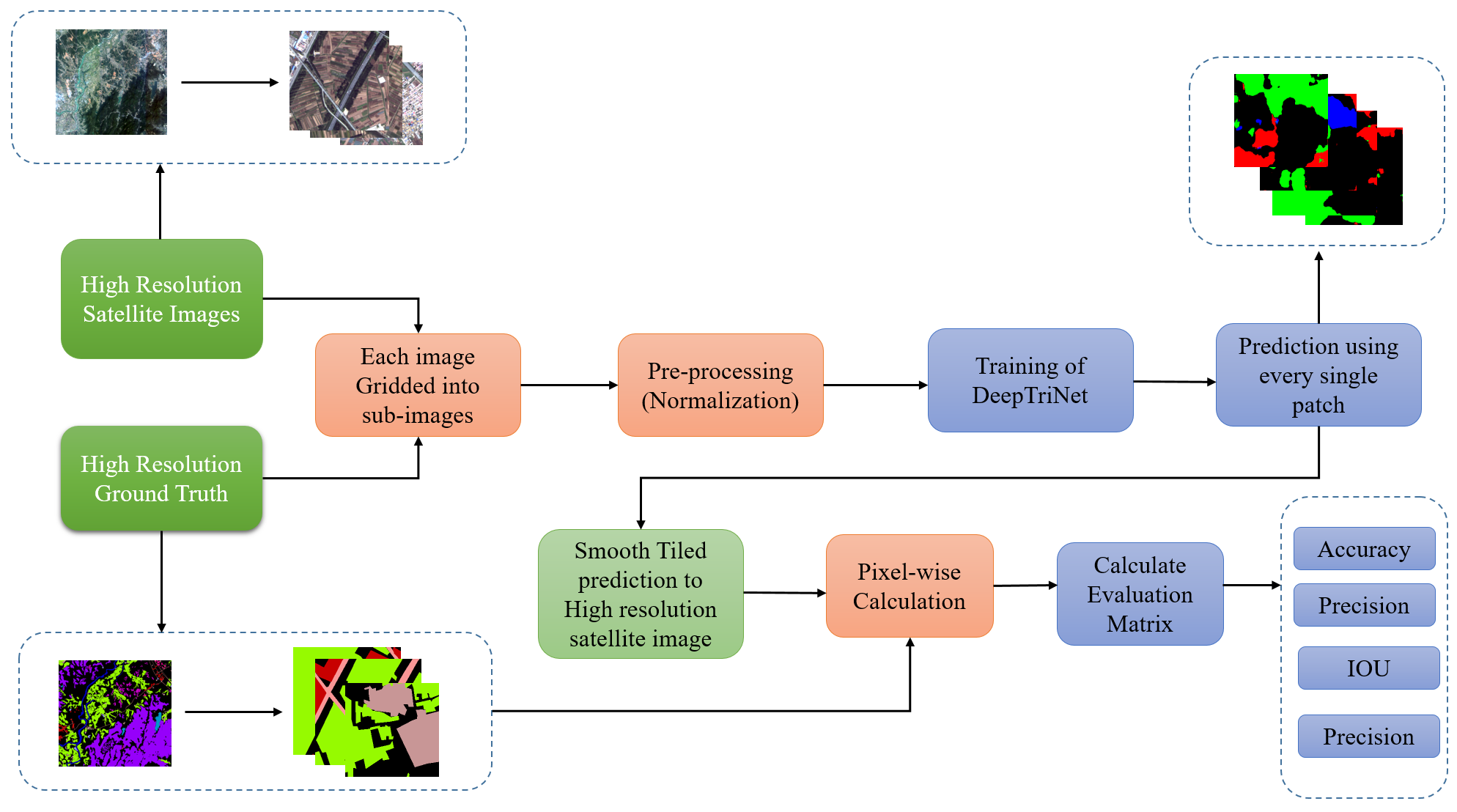}
    \caption{Research Workflow}
    \label{research_workflow}
\end{figure*}
\end{itemize}

\section{Result Analysis}
Our proposed architecture has been evaluated based on the basis of accuracy, precision, recall and IoU.

\begin{table*}[htbp]
	\centering
    \small
	\renewcommand{\arraystretch}{1}
	\caption{Performance Comparison with Existing Work \label{tab:comp}}
	\resizebox{\textwidth}{!}{
        \begin{tabular}{|p{.91cm}|p{2cm}|p{1.9cm}|p{.89cm}|p{1.5cm}|p{2.9cm}|p{5cm}|}
		%\begin{tabular}{cllclll}
			\toprule
			\textbf{Year} & \textbf{Reference} & \textbf{Dataset} & 
			\textbf{No of Clas- ses}  & \textbf{Pre pro- cessing}  &  \textbf{Network(Type)}  & \textbf{Matrices}\\
			\hline
   2022 & Lee~\etal\cite{lee2022comparisons} & LandCover.ai & 4 & grid and patch & DeepLabv3+(CNN) & \makecell[l]{Accuracy: 0.85
				Precision: NA\\ Recall: NA IoU: NA} \\
    		\midrule
      2022 & Lee~\etal\cite{lee2022comparisons} & LandCover.ai & 4 & grid and patch & SegNet(CNN) & \makecell[l]{Accuracy: 0.88
				Precision: NA\\ Recall: NA IoU: NA } \\
    		\hline
				2019 & Yao~\etal~\cite{Yao2019} &  ISPRS & 6 & single image & DeepLabv3+(CNN) & \makecell[l]{Accuracy:  0.91
				Precision: 0.80\\ Recall: NA IoU: NA } \\
			\hline
   2020 & Nayem~\etal~\cite{Nayem2020} & Gaofen-2 & 6 & grid and patch & FCN-8(CNN) & \makecell[l]{Accuracy: 0.91
				Precision: NA\\ Recall: NA IoU: 0.840} \\
    \hline
    2022 & Lee~\etal\cite{lee2022comparisons} & LandCover.ai & 4 & grid and patch & U-net(CNN) & \makecell[l]{Accuracy:0.91
				Precision: NA\\ Recall: NA IoU: NA } \\
    		\hline
      2020 & Onim~\etal~\cite{Onim2020} & Gaofen-2 & 6 &grid and patch & FastFCN(CNN) &  \makecell[l]{Accuracy:  0.93
				Precision:  0.99\\ Recall: 0.98  IoU: 0.97} \\
	    	\hline
    2021 & Kang~\etal~\cite{KANG2021102499} & LandCover.ai & 1 & grid and patch & Multi-scale context extractor with CNN & \makecell[l]{Accuracy: 0.98
				Precision: NA\\ Recall: NA IoU:0.938} \\
    		
			\hline
   
            	\textbf{2023} & \textbf{Ours}  &  \textbf{Gaofen-2} & \textbf{15} & \textbf{grid and patch} &  \makecell[l]{\textbf{DeepTriNet}} & \makecell[l]{ \bf Accuracy: 0.77
				\bf Precision: 0.68\\ \bf Recall: 0.55 \bf IoU: 0.58} 
				\\
    \hline
    \textbf{2023} & \textbf{Ours}  &  \textbf{LandCover .ai} & \textbf{4} &\textbf{grid and patch} &  \makecell[l]{\textbf{DeepTriNet}} & \makecell[l]{ \bf Accuracy: 0.98
				\bf Precision: 0.80\\ \bf Recall: 0.79\\ \bf IoU: 0.80} 
				\\
        	
			\bottomrule
		\end{tabular}
 	}
  \label{performance_table}
\end{table*}
\begin{center}
    \centering
    \includegraphics[width=.95\columnwidth]{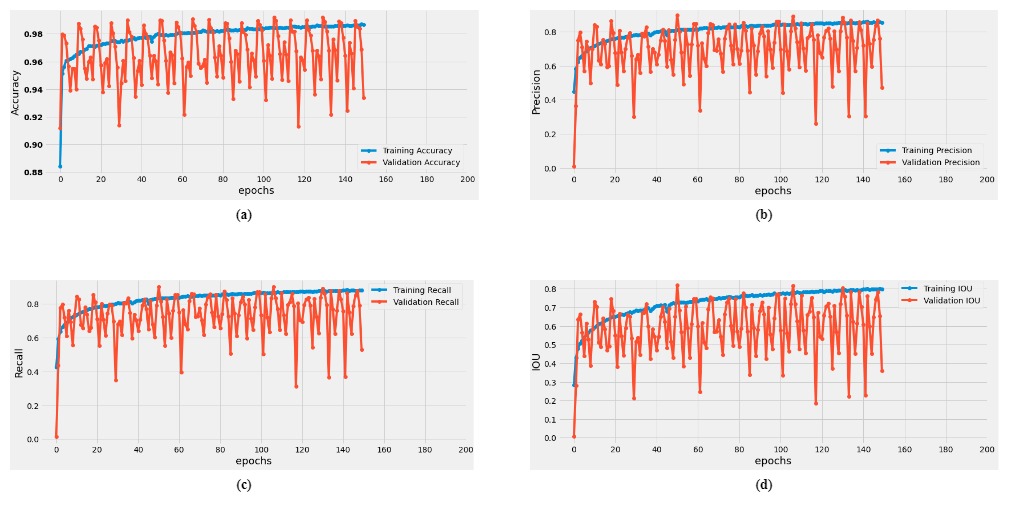}
    \captionof{figure}{Performance History of DeepTriNet on LandCover.ai Dataset}
    \label{history_lc}
\end{center}

\begin{center}
    \centering
    \includegraphics[width=0.95\columnwidth]{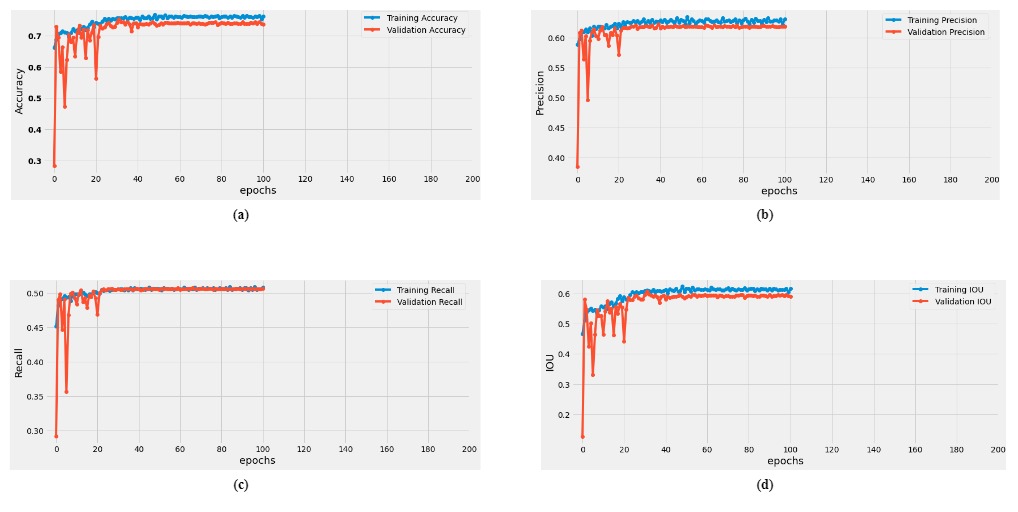}
    \captionof{figure}{Performance History of DeepTriNet on GID-2 Dataset}
    \label{history_gid}
\end{center} 

Fig.\ref{history_lc} and Fig.\ref{history_gid} show the training history of the proposed DeepTriNet model in terms of accuracy, precision, recall and IoU. It can be easily observed from both Fig.\ref{history_lc} and Fig.\ref{history_gid} that there were no overfitting while training the DeepTriNet, rather a perfect fitting curve for all the evaluation parameter has been shown.
In Table.\ref{performance_table} a comparative analysis between the existing literature and our proposed DeepTriNet architecture has been summarized.

\section{Conclusion}
In conclusion, DeepTriNet introduces a self-supervised DeepLabv3+ architecture based on tri-level attention for semantic segmentation of satellite images. The proposed model combines the vanilla DeepLabv3+ architecture with Tri-Level Attention Units (TAUs) and Squeeze-and-Excitation Networks (SENets) to increase segmentation efficiency and precision. 

Experimental findings on the $4$-class Land-Cover.ai dataset demonstrate that DeepTriNet outperforms conventional approaches, except for a method that only considers one class. Also presented are other measures, including accuracy, recall, and the F1 score. The performance of DeepTriNet for $15$-class GID-2 dataset is noteworthy also comparing the data volume and diversity.

Better natural resource management and change detection in rural and urban regions are two uses of this technology. However, this approach has drawbacks, including the demand for a sizable amount of training data and computer power. 

The potential of DeepTriNet in other domains, such as autonomous driving or medical imaging, may be investigated in subsequent research. Overall, the contributions of this model provide a viable method for precise and effective satellite picture segmentation, with potential applications in many different sectors.

\bibliographystyle{style}
\bibliography{reference}
\end{document}